\documentclass{article}
\usepackage{amsmath,epsfig}
\usepackage[preprint]{spconfa4}
\usepackage{amssymb}
\usepackage{mathrsfs}
\usepackage{bm}
\usepackage{multirow}

\copyrightnotice{978-1-7281-1331-9/20/\$31.00~\copyright 2020 IEEE}

\let\OLDthebibliography\thebibliography
\renewcommand\thebibliography[1]{
  \OLDthebibliography{#1}
  \setlength{\parskip}{0pt}
  \setlength{\itemsep}{0pt plus 0.3ex}
}

\begin{document}\sloppy

\def\x{{\mathbf x}}
\def\L{{\cal L}}

\title{GRAPH-BASED KINSHIP REASONING NETWORK}
%

\name{Wanhua Li$^{1,2,3}$, Yingqiang Zhang$^{4}$, Kangchen Lv$^{1}$, Jiwen Lu$^{1,2,3,*}$, Jianjiang Feng$^{1,2,3}$, Jie Zhou$^{1,2,3}$\thanks{This work was supported in part by the National Key Research and Development Program of China under Grant 2017YFA0700802, in part by the National Natural Science Foundation of China under Grant 61822603, Grant U1813218, Grant U1713214, and Grant 61672306, in part by Beijing Academy of Artificial Intelligence (BAAI), in part by the Shenzhen Fundamental Research Fund (Subject Arrangement) under Grant JCYJ20170412170602564, and in part by Tsinghua University Initiative Scientific Research Program.}\thanks{$^{*}$Corresponding author.}}
\address{$^{1}$Department of Automation, Tsinghua University, China\protect\\
$^{2}$State Key Lab of Intelligent Technologies and Systems, China\protect\\
$^{3}$Beijing National Research Center for Information Science and Technology, China\protect\\
$^{4}$Central Media Technology Institute, Huawei Technology Co., Ltd., China\protect\\
\{li-wh17, lkc17\}@mails.tsinghua.edu.cn,  \qquad zhangyingqiang1@huawei.com,\protect\\
\{lujiwen, jfeng, jzhou\}@tsinghua.edu.cn}

\maketitle

\begin{abstract}
In this paper, we propose a graph-based kinship reasoning (GKR) network for kinship verification, which aims to effectively perform relational reasoning on the extracted features of an image pair. Unlike most existing methods which mainly focus on how to learn discriminative features, our method considers how to compare and fuse the extracted feature pair to reason about the kin relations. The proposed GKR constructs a star graph called kinship relational graph where each peripheral node represents the information comparison in one feature dimension and the central node is used as a bridge for information communication among peripheral nodes. Then the GKR performs relational reasoning on this graph with recursive message passing. Extensive experimental results on the KinFaceW-I and KinFaceW-II datasets show that the proposed GKR outperforms the state-of-the-art methods.
\end{abstract}
\section{Introduction}
\label{sec:intro}

Some research~\cite{dal2010lateralization} in biology finds that  human facial appearance contains important kin related information. Inspired by this finding, many methods~\cite{lu2013neighborhood,yan2014discriminative} have been proposed for kinship recognition from facial images. The goal of kinship verification is to determine whether or not a kin relation exists for a given pair of facial images. Kinship verification has attracted increasing attention in the computer vision community due to its broad applications such as automatic album organization~\cite{zhou2012gabor}, missing children searching~\cite{lu2013neighborhood},  social media-based analysis~\cite{dehghan2014look}, and children adoptions~\cite{BMVC2015_148}.

Although a variety of efforts~\cite{lu2013neighborhood,yan2014discriminative,BMVC2015_148} have been devoted to kinship verification, it is still far from ready to be deployed for any real-world uses. There are several challenges preventing the development of kinship recognition.
First, as other face-related tasks~\cite{liu2017sphereface,li2019bridgenet}, facial kinship verification is also confronted with a large variation of the pose, scale, and illumination, which makes learning discriminative features quite challenging. Second, unlike face verification which investigates the relations between different images of an entity, kinship verification has to discover the hidden similarity inherited by genetic relations between different identities, which naturally leads to a much larger appearance gap of intra-class samples, especially when there are significant gender differences and age gaps in kinship verification.

\begin{figure}[t]
  \centering
  \includegraphics[width=1.0\linewidth]{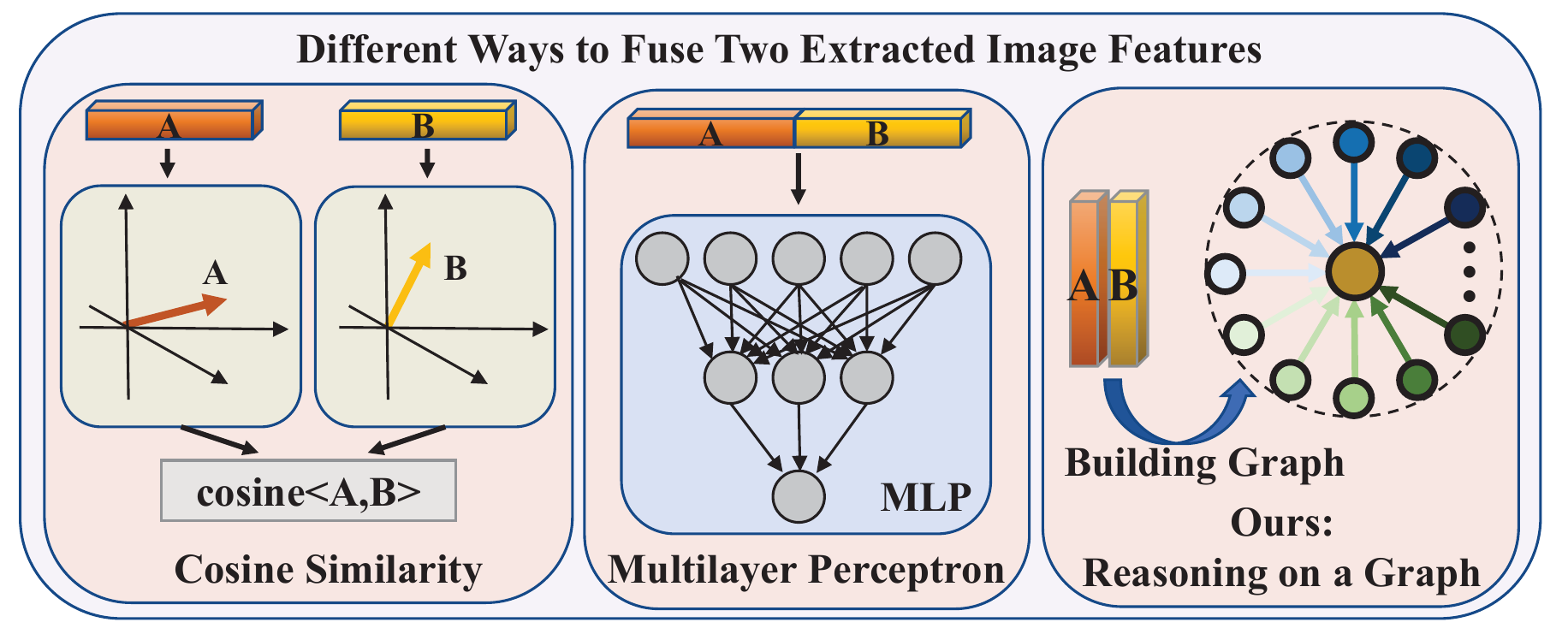}
  \caption{ Key differences between our method and other methods. Most existing methods usually apply a similarity metric such as cosine similarity, or a multilayer perceptron to the extracted image features, which couldn't fully exploit the hidden relations. On the other hand, our method builds a kinship relational graph and performs relational reasoning on this graph.}
  \label{fig:introduction}
\end{figure}

Many methods have been proposed to address these challenges over the past few years. Most of them pay attention to learning discriminative features for each facial image of a paired sample. For example, Lu \emph{et al.}~\cite{lu2013neighborhood} proposed the NRML metric to pull intra-class samples as close as possible and push interclass samples in a neighborhood as far as possible in the learned feature space. Nevertheless, these approaches usually apply a similarity metric~\cite{lu2017discriminative} or a multilayer perceptron (MLP)~\cite{BMVC2015_148} to the extracted features to obtain the probability of kinship between two facial images, which couldn't fully exploit the genetic relations of two features.

In this paper, we focus on the aspect of how to compare and fuse the two extracted features of a paired sample to reason about the genetic relations. We hypothesize that when people reason about the kinship relations, they usually first compare the genetically related attributes of two individuals, such as cheekbone shape, eye color, and nose size, and then make comprehensive judgments based on these comparison results. For two given features of a paired sample, we consider that each dimension of the feature encodes one kind of kin related information. Therefore, we explicitly model the reasoning process of humans by comparing the features for each dimension and then fusing them. More specifically, we build a star graph named kinship relational graph for two features to perform relational reasoning, where each peripheral node models one dimension of features and the central node is utilized as a bridge of communication. We further propose a graph-based kinship reasoning (GKR) network on this graph to effectively exploit the hidden kin relations of extracted features. The key differences of our method and most existing methods are visualized in Figure \ref{fig:introduction}. We validate the proposed GKR for kinship verification on two benchmarks: KinFaceW-I~\cite{lu2013neighborhood} and KinFaceW-II~\cite{lu2013neighborhood} datasets, and the results illustrate that our method outperforms the state-of-the-art approaches.

\section{RELATED WORK}

\textbf{Kinship Verification:} In the past few years, many papers~\cite{yan2014discriminative,BMVC2015_148,fang2010towards,zhou2019learning} have been published for kinship verification and most of them focus on extracting discriminative features for each image, which can be divided into three categories: hand-crafted methods, distance metric-based methods, and deep learning-based methods.

Hand-crafted methods require researchers to design the feature extractors by hand. For example, as one of the earliest works, Fang \emph{et al.}~\cite{fang2010towards} proposed extracting color, facial parts, facial distances, and gradient histograms as the features for classification. Zhou \emph{et al.}~\cite{zhou2012gabor} further presented a Gabor-based gradient orientation pyramid (GGOP) feature representation method to make better of multiple feature information. Distance metric-based methods ~\cite{zhao2018learning,mahpod2018kinship} are the most popular methods for kinship verification, which aim to learn a distance metric such that the distance between positive face pairs is reduced and that of negative samples is enlarged. Yan \emph{et al.}~\cite{yan2014discriminative} first extracted multiple features with different descriptors and then learned multiple distance metrics to exploit complementary and discriminative information. A discriminative deep metric learning method was introduced in ~\cite{lu2017discriminative}, which learned a set of hierarchical nonlinear transformations with deep neural networks.
Zhou \emph{et al.}~\cite{zhou2019learning} explicitly considered the discrepancy of cross-generation and a kinship metric learning method with a coupled deep neural network was proposed to improve the performance.
Recent years have witnessed the great success of deep learning. However, few deep learning-based works have been done for kinship verification. Zhang \emph{et al.}~\cite{BMVC2015_148} were the first attempt for kinship verification with deep CNNs and demonstrated the effectiveness of their method. Hamdi~\cite{dibeklioglu2017visual} further studied video-based kinship verification with deep learning. All these methods only focus on learning good feature representations, which ignore how to 
reason about the kin relations with extracted embeddings.

\noindent
\textbf{Graph Neural Networks:} A variety of practical applications deal with the complex non-Euclidean structure of graph data. Graph neural networks (GNNs) are proposed to handle these kinds of data, which learn features on graphs. Li \emph{et al.}~\cite{li2015gated} proposed gated graph neural networks (GG-NNs) with gated recurrent units, which could be trained with modern optimization techniques. Motivated by the success of convolutions on the image data, Kipf and Welling introduced the graph convolutional networks (GCNs)~\cite{kipf2016semi} by applying the convolutional architecture on graph-structured data. A layer-wise propagation rule was utilized in GCNs and both local graph structure and node features were encoded for the task of semi-supervised learning.  Petar \emph{et al.}~\cite{velivckovic2017graph} further proposed the graph attention networks (GATs) to assign different weights to different nodes in a self-attentional way. The generated weights didn't require any prior knowledge of the graph structure. GATs were computationally efficient and had a larger model capacity due to the attention mechanism. GNNs have proven to be a good tool for relational reasoning. For example, Sun \emph{et al.}~\cite{sun2019relational} constructed a recurrent graph to jointly model the temporal and spatial interactions among different individuals with GNNs for action forecasting.

\section{PROPOSED APPROACH}
In this section, we first present the problem formulation. Then we illustrate the details of the kinship relational graph building process. Lastly, we introduce the proposed graph-based kinship reasoning (GKR) network.

\begin{figure*}[t]
  \centering
  \includegraphics[width=1.0\linewidth]{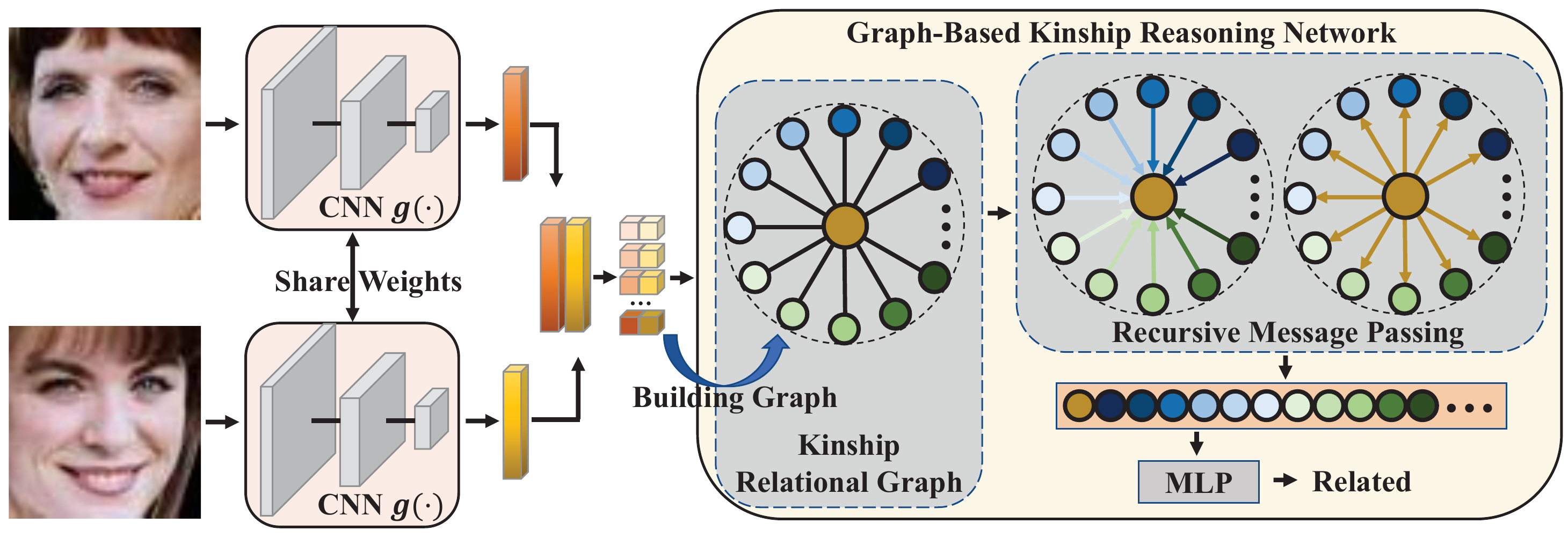}
  \caption{ An overall framework of our proposed approach. For a given image pair, we first send them to the same CNN to extracted features. Then we build a kinship relational graph with these two features, where each surrounding node is initialized with the values of two features in one dimension. Our method employs a recursive message passing scheme to perform relational reasoning on this graph. In the end, all node features are concatenated and sent to an MLP to generate the prediction.}
  \label{fig:framework}
\end{figure*}

\subsection{Problem Formulation}
We use $\mathcal{P} = \{(\bm{x}_i,\bm{y}_i)|i=1,2,...,N\}$ to denote the training set of paired images with kin relations, where $\bm{x}_i$ and $\bm{y}_i$ are the parent image and child image, respectively, and $N$ is the total number of the positive training set. Therefore, the negative training set is built as $\mathcal{N} = \{(\bm{x}_i,\bm{y}_j)| i,j = 1,2,...,N, i \neq j\}$, where each parent image and each unrelated child image form a negative sample. However, the size of the negative training set is much larger than that of positive training set given that $|\mathcal{P}| = N$ and $|\mathcal{N}| = N(N-1)$. So we randomly select negative samples from the set $\mathcal{N}$ to build a balanced negative training set $\mathcal{N}'$ such that $|\mathcal{N}'| = N$. Then the whole training set $\mathcal{D}$ is constructed with the union of the positive training set and negative training set:  $\mathcal{D} = \mathcal{P} \bigcup \mathcal{N}'$.

The goal of kinship verification can be formulated as learning a mapping function, where the input is a paired sample $(\bm{x}_i,\bm{y}_j)$ and the output is the probability of $i = j$. Most existing methods aim to learn a good feature extractor $g(\cdot)$. Hand-crafted methods usually design shallow features by hand to implement $g(\cdot)$, whereas deep learning-based methods usually learn a deep neural network as the extractor $g(\cdot)$. Metric learning-based methods usually first use hand-crafted features or deep features as the initial sample features $(g'(\bm{x}_i),g'(\bm{y}_j))$, and then learn a distance metric:
\begin{equation}
d(\bm{x}_i,\bm{y}_j) = \sqrt{d'(\bm{x}_i,\bm{y}_j)^T \bm{W} \bm{W}^T d'(\bm{x}_i,\bm{y}_j)},
\label{equ:metric}
\end{equation}
where $d'(\bm{x}_i,\bm{y}_j) = g'(\bm{x}_i) - g'(\bm{y}_j)$. In the end, we obtain the projected features $g(\bm{x}_i) = \bm{W}^T g'(\bm{x}_i) \in \mathbb{R}^{D}$ and $g(\bm{y}_j) = \bm{W}^T g'(\bm{y}_j) \in \mathbb{R}^{D}$, where $D$ denotes the feature dimension.

Having obtained the features $(g(\bm{x}_i),g(\bm{y}_j)) \in (\mathbb{R}^{D},\mathbb{R}^{D})$, we still need to learn a mapping function $f(\cdot)$ to map the features $(g(\bm{x}_i),g(\bm{y}_j))$ to a probability of kin relation between $\bm{x}_i$ and $\bm{y}_j$. Most current methods mainly focus on the feature extractor $g(\cdot)$ and usually neglect the design of $f(\cdot)$. One choice is to simply concatenate two features and send them to a multilayer perceptron (MLP):
\begin{equation}
f(g(\bm{x}_i),g(\bm{y}_j)) = {\rm MLP}([g(\bm{x}_i)||g(\bm{y}_j)]),
\label{equ:fuse_MLP}
\end{equation}
where $||$ represents the concatenation operation. Another commonly used way is to calculate the cosine similarity of two features:
\begin{equation}
f(g(\bm{x}_i),g(\bm{y}_j)) = \frac{g(\bm{x}_i)^Tg(\bm{y}_j)}{\Vert g(\bm{x}_i)\Vert  \Vert  g(\bm{y}_j) \Vert  }.
\label{equ:fuse_cos}
\end{equation}
Both methods can't fully exploit the relations of two features. In this paper, we aim to design a new $f(\cdot)$ to effectively perform relational reasoning on the two extracted features.

\subsection{Building a Kinship Relational Graph}
In recent years, deep CNNs have achieved great success in many computer vision tasks, such as image classification, object detection, and scene understanding, which demonstrates their superior ability for feature representation. Therefore, we utilize a deep CNN as the feature extractor $g(\cdot)$ in this paper.

Having obtained the deeply learned sample features $(g(\bm{x}_i),g(\bm{y}_j))$, we consider how to perform relational reasoning on them. To achieve this, we first observe how humans reason about kin relations. As the genetic traits are usually exhibited by facial characteristics, humans reason about the kin relations by comparing the genetically related attributes to discover the hidden similarity. For example, if we find that the persons on two facial images have the same eye color and similar cheekbones, the probability that they are related will be higher. After comparing a variety of informative facial attributes of two persons, humans make the final decision by combining and analyzing all the information.

We explicitly model the above reasoning process by constructing a kinship relational graph and performing relational reasoning on this graph. We consider that each dimension of the extracted features encodes one kind of genetic information and we can reason about the kin relations by comparing and fusing all the genetic information. Since we use the same CNN to extract features for two images, the values of two features in the same dimension represent the comparison of one kind of kinship related information encoded in that dimension. We use one node in the kinship relational graph to denote the comparison of one feature dimension, then we have $D$ nodes which describe the comparisons in all dimensions. To fuse these comparisons, we need to define the interactions of these $D$ nodes. One intuitive way is to connect all the nodes given that any two nodes may have a relation. However, such a graph greatly increases the computational complexity of subsequent operations. Therefore, we create a super node that is connected to all other nodes while all other nodes are only connected to the super node. The super node is also the central node of the star-structured kinship relational graph, which plays an important role in the interaction and information communication of $D$ surrounding nodes.  In this way, we build the kinship relational graph and will elaborate on the reasoning process with the proposed graph-based kinship reasoning network in the following subsection.

\subsection{Reasoning on the Kinship Relational Graph}

Having built the kinship relational graph, we consider how to perform relational reasoning on this graph. Recently, graph neural networks (GNNs) have attracted increasing attention for representation learning of graphs. Generally speaking, GNNs employ a recursive message-passing scheme, where each node aggregates the messages sent by its neighbors to update its feature. We follow this scheme and propose the graph-based kinship reasoning (GKR) network to perform relational reasoning on the kinship relational graph.

Formally, Let $\mathcal{G} = (\mathcal{V},\mathcal{E})$ denote the kinship relational graph with the node set  $\mathcal{V}$ and the edge set $\mathcal{E}$. Each node in the graph has a feature vector and we have $\mathcal{V} = \{\bm{h_c} \} \bigcup \{\bm{h_d}|d=1,2,...,D\}$, where $\bm{h_c}$ represents the feature vector of the central node and $\bm{h_d}$ is that of $d^{th}$ surrounding node. The edge set of this graph is formulated as $\mathcal{E} = \{e_{cd} | d = 1,2,...,D\}$, where $e_{cd}$ denotes the edge between node $\bm{h_c}$ and $\bm{h_d}$. The proposed GKR propagates messages according to the graph structure defined by $\mathcal{E}$ and the aggregated messages are utilized to update the node features. As mentioned above, we set the initial node features as the values of two extracted image features in one dimension. Mathematically, the initial node features are set as follows:
\begin{equation}
\bm{h_d^0} = [g_d(\bm{x}_i) || g_d(\bm{y}_j)],
\label{equ:ini_node}
\end{equation}
where $\bm{h_d^0} \in \mathbb{R}^{2}$ denotes the initial feature of $d^{th}$ node, $g_d(\bm{x}_i)$ and $g_d(\bm{y}_j)$ represent the values in the $d^{th}$ dimension of features  $g(\bm{x}_i)$ and $g(\bm{y}_j)$, respectively. In this way, each node encodes one kind of kinship related information.

The proposed GKR consists of $K$ layers where each layer represents one time step of the message passing phase. The $k^{th} (1 \leq k \leq K)$ layer transforms the node features $\bm{h_c^{k-1}},\bm{h_1^{k-1}},\bm{h_2^{k-1}},...,\bm{h_D^{k-1}} \in \mathbb{R}^{F_{k-1}}$ into $\bm{h_c^{k}},\bm{h_1^{k}},\bm{h_2^{k}},...,\bm{h_D^{k}} \in \mathbb{R}^{F_{k}}$ with message passing to perform relational reasoning, where $\mathbb{R}^{F_{k-1}}$ and $\mathbb{R}^{F_{k}}$ represent the corresponding feature dimensions. Having obtained the node features of the $(k-1)^{th}$ layer, we first generate the message of each node which is going to be sent out in the following message passing process. The message of the surrounding node is generated following:
\begin{equation}
\bm{m_d^{k}} = {\rm ReLU}(\bm{W}_{mess}^T \bm{h_d^{k-1}}), d= 1,2,...,D
\label{equ:mess1}
\end{equation}
where $\bm{W}_{mess} \in \mathbb{R}^{F_{k-1} \times F_{k}}$ is employed to transform the node features into messages. We apply the same operation  for the central node with the same parameter $\bm{W}_{mess}$:
\begin{equation}
\bm{m_c^{k}} = {\rm ReLU}(\bm{W}_{mess}^T \bm{h_c^{k-1}}).
\label{equ:mess2}
\end{equation}

With these messages, we propagate and aggregate them according to the graph structure. Then we update the node features with the aggregated messages. For the peripheral nodes, since the central node is the only neighbor node, the aggregation is implemented by concatenating the message of the central node and its own message. Then we use the aggregated messages to update the node feature as follows:
\begin{equation}
\bm{h_d^{k}} = {\rm ReLU}(\bm{W}_{peri}^T [\bm{m_d^{k}} || \bm{m_c^{k}} ] ), d= 1,2,...,D
\label{equ:update1}
\end{equation}
where $\bm{W}_{peri} \in \mathbb{R}^{2F_{k} \times F_{k}}$ is used to fuse all information to generate the new feature vector. For the central node, we first aggregate all the incoming messages:
\begin{equation}
\bm{a^{k}} = {\rm AGGREGATE}(\{ \bm{m_d^k} | d = 1,2,...,D\}),
\label{equ:agg}
\end{equation}
where the function AGGREGATE$(\cdot)$ is implemented by a pooling operation. Then the feature of the central node is updated as follows:
\begin{equation}
\bm{h_c^{k}} = {\rm ReLU}(\bm{W}_{cen}^T [\bm{m_c^{k}} || \bm{a^{k}} ] ),
\label{equ:update2}
\end{equation}
where $\bm{W}_{cen} \in \mathbb{R}^{2F_{k} \times F_{k}}$ is utilized to update the feature of the central node. In this way, we obtain the updated features $\bm{h_c^{k}},\bm{h_1^{k}},\bm{h_2^{k}},...,\bm{h_D^{k}}$ by message passing.

We repeat the above process for $K$ times and have the final node feature vectors: $\bm{h_c^{K}},\bm{h_1^{K}},\bm{h_2^{K}},...,\bm{h_D^{K}} \in \mathbb{R}^{F_{K}}$. To make the final decision, we first combine all these features and send them to an MLP, which outputs a scalar value. Therefore, the mapping function $f(\cdot)$ of our proposed method can be formulated as:
\begin{equation}
f(g(\bm{x}_i),g(\bm{y}_j)) = {\rm MLP}([\bm{h_c^{K}} || \bm{h_1^{K}} || \bm{h_2^{K}} || ... || \bm{h_D^{K}} ]).
\label{equ:res}
\end{equation}
Lastly, we obtain the probability of kin relation between $\bm{x_i}$ and $\bm{y_j}$ by applying a sigmoid function to the scalar value $f(g(\bm{x}_i),g(\bm{y}_j))$.

Note that the proposed GKR and the feature extractor network $g(\cdot)$ are trained end-to-end. We employ the binary cross-entropy loss as the objective function:
\begin{equation}
\begin{aligned}
\mathcal{L} =& - \frac{1}{N} \sum_{(\bm{x},\bm{y}) \in \mathcal{P}} \log(f(g(\bm{x}),g(\bm{y})))  \\
 & - \frac{1}{N} \sum_{(\bm{x},\bm{y}) \in \mathcal{N}'}  \log(1 - f(g(\bm{x}),g(\bm{y}))).
\end{aligned}
\label{equ:loss}
\end{equation}

In this way, our method is optimized in a class-balanced setting. Lastly, we depict the above pipeline in Figure \ref{fig:framework}.

\begin{table*}[thbp]
\caption{Verification accuracy of different methods on KinFaceW-I and KinFaceW-II datasets.}
\vspace {0.2cm}
\label{table:stoa}
\centering
\begin{tabular}{|l|ccccc|ccccc|}
\hline
\multirow{2}*{Method} & \multicolumn{5}{|c|}{KinFaceW-I} & \multicolumn{5}{c|}{KinFaceW-II}\\
\cline{2-11}
~ & F-S & F-D & M-S & M-D & Mean & F-S & F-D & M-S & M-D & Mean\\
\hline \hline
MNRML~\cite{lu2013neighborhood} & 72.5\% & 66.5\% & 66.2\% & 72.0\% & 69.9\% & 76.9\% & 74.3\% & 77.4\% & 77.6\% & 76.5\% \\
DMML~\cite{yan2014discriminative} & 74.5\% & 69.5\% & 69.5\% & 75.5\% &  72.3\% & 78.5\% & 76.5\% & 78.5\% & 79.5\% & 78.3\% \\
CNN-Basic~\cite{BMVC2015_148} & 75.7\% & 70.8\% & 73.4\% & 79.4\% & 74.8\% & 84.9\% & 79.6\% & 88.3\% & 88.5\% & 85.3\% \\
CNN-Point~\cite{BMVC2015_148} & 76.1\% & 71.8\% & 78.0\% & 84.1\% & 77.5\% & 89.4\% & 81.9\% & 89.9\% & 92.4\% & 88.4\% \\
D-CBFD~\cite{yan2019learning} & 79.0\% & \textbf{74.2\%} & 75.4\% & 77.3\% & 78.5\% & 81.0\% & 76.2\% & 77.4\% & 79.3\% & 78.5\% \\
WGEML~\cite{liang2018weighted} & 78.5\% & 73.9\% & \textbf{80.6\%} & 81.9\% & 78.7\% & 88.6\% & 77.4\% & 83.4\% & 81.6\% & 82.8\% \\
\hline
GKR & \textbf{79.5\%} & 73.2\% & 78.0\% & \textbf{86.2\%} & \textbf{79.2\%} & \textbf{90.8\%} & \textbf{86.0\%} & \textbf{91.2\%} & \textbf{94.4\%} & \textbf{90.6\%} \\
\hline
\end{tabular}
\end{table*}

\section{EXPERIMENTS}

In this section, we conducted extensive experiments on two widely-used kinship  verification datasets to illustrate the effectiveness of the proposed GKR.

\subsection{Datasets and Implementation Details}
We employ two widely-used databases: KinFaceW-I~\cite{lu2013neighborhood} and KinFaceW-II~\cite{lu2013neighborhood} for evaluation, which are collected from the internet. Four different types of kinship relations are considered in these two datasets: Father-Daughter (F-D), Father-Son (F-S), Mother-Daughter (M-D), and Mother-Son (M-S). There are 156, 134, 116 and 127 pairs of facial images for these four relations in KinFaceW-I, respectively while KinFaceW-II contains 250 pairs of parent-child facial images for each kin relation. The main difference between these two databases is that each image pair with kin relation in KinFaceW-I comes from different photos whereas that in KinFaceW-II is collected from the same photo.

We employed the ResNet-18 as the feature extractor network $g(\cdot)$, which was initialized with the ImageNet pre-trained weights. Naturally, the dimension $D$ of the extracted image features was equal to 512. Since both databases are relatively small, data augmentation is a crucial step to improve performance. We performed data augmentation by first resizing the facial images into 73 $\times$ 73 pixels and then random cropping a 64 $\times$ 64 patch. Following the design choice of most GNNs methods~\cite{kipf2016semi}, we used a two-layer ($K=2$) GKR and let $F_1 = 512, F_2 = 4$. Adam optimizer was utilized with a learning rate of 0.0005. The batch size was set to 16 and 32 for  KinFaceW-I and  KinFaceW-II, respectively given that the size of the KinFaceW-I database is only about half the size of the KinFaceW-II database. For a fair comparison, we performed the five-fold cross-validation following the standard protocol provided in~\cite{lu2013neighborhood}.

\begin{table}[thbp]
\caption{Mean verification accuracy of different initialization strategies of the central node. }
\renewcommand\tabcolsep{4.5pt}
\vspace {0.1cm}
\label{table:Initial}
\centering
\begin{tabular}{|l|ccccc|}
\hline
\multirow{2}*{Dataset} & \multicolumn{5}{|c|}{Initialization} \\
\cline{2-6}
~ & Mean & Max & 0 & 0.5 & 1 \\
\hline
KinFaceW-I & 77.4\% &  73.5\% &  77.5\% &  \textbf{79.2\%}&  78.1\% \\
KinFaceW-II & 79.1\% &  80.6\% &  79.5\% &  \textbf{90.6\%} &  87.5\%\\
\hline
\end{tabular}
\end{table}

\begin{table}[thbp]
\caption{Comparisons of different pooling operations for AGGREGATE$(\cdot)$ (KFW-I for KinFaceW-I and KFW-II for  KinFaceW-II ). }
\renewcommand\tabcolsep{3.5pt}
\vspace {0.1cm}
\label{table:aggrate}
\centering
\begin{tabular}{|l|c|ccccc|}
\hline
Dataset & Pool & F-S & F-D & M-S & M-D & Mean \\
\hline
\multirow{2}*{KFW-I} & Mean &  78.2\% &  \textbf{76.8\%}  & 73.8\% & \textbf{86.4\%} & 78.8\% \\
 ~ & Max & \textbf{79.5\%} &  73.2\%  & \textbf{78.0\%} & 86.2\% & \textbf{79.2\%} \\
\hline
\multirow{2}*{KFW-II} & Mean & 90.2\% & \textbf{87.0\%} & 90.3\% & 92.5\% & 90.0\% \\
~ & Max & \textbf{90.8\%} & 86.0\% & \textbf{91.2\%} & \textbf{94.4\%} & \textbf{90.6\%} \\
\hline
\end{tabular}
\end{table}

\subsection{Comparison with the State-of-the-Art Methods}
 We first compare our GKR with several state-of-the-art methods including metric-learning based methods and deep learning-based methods. Table \ref{table:stoa} shows the comparison results on KinFaceW-I and KinFaceW-II datasets. We observe that our method achieves an average verification accuracy of 79.2\% on KinFaceW-I and that of 90.6\% on KinFaceW-II, which outperforms state-of-the-art methods. Some early metric learning-based methods, such as MNRML~\cite{lu2013neighborhood} and DMML~\cite{yan2014discriminative} learn the proposed metric with hand-crafted features, which leads to unsatisfactory results. The method of WGEML~\cite{liang2018weighted} achieves state-of-the-art results with deep features, which demonstrates the superiority of deep learning. Compared with WGEML, our method improves the mean accuracy by 0.5\% and 7.8\% on KinFaceW-I and KinFaceW-II,  respectively, which shows the superior relational reasoning ability of the proposed GKR. Zhang \emph{et al.}~\cite{BMVC2015_148} propose the CNN-Basic and CNN-Point, which directly learn deep neural networks for kinship verification to exploit the power of deep learning. Our method, which is also a deep learning-based method, outperforms the CNN-Point by 1.7\% and 2.2\% on KinFaceW-I and KinFaceW-II,  respectively. Note that the CNN-Point contains 10 CNN backbones whereas our approach only employs one CNN backbone, which further illustrates the effectiveness of the proposed GKR.

\subsection{Ablation Study}
To investigate the influence of individual design choices and validate the effectiveness of the proposed GKR, we further conducted ablation experiments in this subsection.

\textbf{Initialization of the Central Node:} The initialization of the central node is an important design choice given that the central node is the bridge of the kinship relational graph. One strategy is to aggregate the initial values of all other nodes by mean or max pooling. Another way is to initialize the central node with constant values, such as 0, 0.5, and 1. The results are listed in Table \ref{table:Initial} and we see that the initialization with constant value 0.5 gives the best performance, which is employed in the following experiments.

\textbf{Pooling Operations of AGGREGATE$\bm{(\cdot)}$:} Two different pooling operations: max-pooling and mean-pooling are considered to implement the function AGGREGATE$(\cdot)$. Table \ref{table:aggrate} tabulates the verification accuracy of these two pooling operations. We observe that the max-pooling achieves better results, perhaps it can better select more important information while the mean-pooling treats all messages equally.

\begin{table}[t]
\caption{Comparisons of different design choices for $f(\cdot)$. }
\renewcommand\tabcolsep{3.5pt}
\vspace {0.1cm}
\label{table:abla}
\centering
\begin{tabular}{|l|c|ccccc|}
\hline
Dataset & $f(\cdot)$ & F-S & F-D & M-S & M-D & Mean \\
\hline
\multirow{3}*{KFW-I} & Cos & 60.6\% &  66.4\%  & 64.2\% & 70.5\% & 65.4\% \\
 ~ & MLP & 70.5\% &  71.5\%  & 71.2\% & 79.2\% & 73.1\% \\
  ~ & ours & \textbf{79.5\%} &  \textbf{73.2\%}  & \textbf{78.0\%} & \textbf{86.2\%} & \textbf{79.2\%} \\
\hline
\multirow{3}*{KFW-II} & Cos & 78.2\% & 73.8\% & 76.6\% & 81\% & 77.4\% \\
~ & MLP & 80.6\% & 82.0\% & 80.8\% & 79.4\% & 80.7\% \\
~ & ours & \textbf{90.8\%} & \textbf{86.0\%} & \textbf{91.2\%} & \textbf{94.4\%} & \textbf{90.6\%} \\
\hline
\end{tabular}
\vspace{-0.3cm}
\end{table}

\textbf{Mapping Function $\bm{f(\cdot)}$:} To validate the effectiveness of our proposed GKR, we compare it with other widely used design choices for $f(\cdot)$: MLP as formulated in \eqref{equ:fuse_MLP} and cosine similarity as formulated in \eqref{equ:fuse_cos}. For a fair comparison, all of them employ the ResNet-18 to extract image features. Table \ref{table:abla} shows the results on two datasets. We see that our method outperforms the MLP and cosine similarity by a large margin on both databases, which demonstrates that our method can better exploit the relations of two extracted features and perform relational reasoning with the kinship relational graph.

\section{CONCLUSION}
In this paper, we have proposed a graph-based kinship reasoning network to effectively exploit the generic relations of two features of a sample. Different from other methods, the proposed GKR focuses on how to compare and fuse the two extracted features to perform relational reasoning. Our method first builds a kinship relational graph for two extracted features and then perform relational reasoning on this graph with message passing.
Extensive experimental results on KinFaceW-I and KinFaceW-II databases demonstrate the effectiveness of our approach.

\bibliographystyle{IEEEbib}
\small
\bibliography{icme2020template}

\end{document}